\documentclass[10pt,twocolumn,letterpaper]{article}

\usepackage[pagenumbers]{cvpr}      
\definecolor{cvprblue}{rgb}{0.21,0.49,0.74}
\usepackage[pagebackref,breaklinks,colorlinks,allcolors=cvprblue]{hyperref}
\usepackage{multirow}
\usepackage{amsmath}
\usepackage{amssymb}
\usepackage{amsthm}
\def\paperID{1789} 
\def\confName{CVPR}
\def\confYear{2026}
\usepackage{graphicx}
\usepackage{xcolor}
\newcommand{\best}[1]{\colorbox[HTML]{ffc5c5}{\textbf{#1}}}
\newcommand{\secbest}[1]{\colorbox[HTML]{ffebd8}{\text{#1}}}
\usepackage{float}
\usepackage{makecell}
\title{REArtGS++: Generalizable Articulation Reconstruction with Temporal Geometry Constraint via Planar Gaussian Splatting}
\author{
Di Wu$^{1,2,4}$\thanks{Work done as an intern at Tencent RoboticsX} , Liu Liu $^{3}$\thanks{Corresponding author} ,
Anran Huang$^{3}$, Yuyan Liu$^{3}$, Qiaojun Yu$^{4,5}$, Shaofan Liu$^{3}$, \\
Liangtu Song$^{1}$  , Cewu Lu$^{5}$ \\
1 \textit{Hefei Institutes of Physical Science, Chinese Academy of Sciences}
 \\
2 \textit{University of Science and Technology of China} \\
3 \textit{Hefei University of Technology} \\
4 \textit{RoboticsX, Tencent} \\
5 \textit{Shanghai Jiao Tong University}\\
Email: \texttt{wdcs@mail.ustc.edu.cn, liuliu@hfut.edu.cn} 
}

\begin{document}

\maketitle

\begin{abstract}
Articulated objects are pervasive in daily environments, such as drawers and refrigerators. Towards their part-level surface reconstruction and joint parameter estimation, REArtGS~\cite{wu2025reartgs} introduces a category-agnostic approach using multi-view RGB images at two different states. However, we observe that REArtGS still struggles with screw-joint or multi-part objects and lacks geometric constraints for unseen states. In this paper, we propose REArtGS++, a novel method towards generalizable articulated object reconstruction with temporal geometry constraint and planar Gaussian splatting. We first model a decoupled screw motion for each joint without type prior, and jointly optimize part-aware Gaussians with joint parameters through part motion blending. To introduce time-continuous geometric constraint for articulated modeling, we encourage Gaussians to be planar and propose a temporally consistent regularization between planar normal and depth through Taylor first-order expansion. Extensive experiments on both synthetic and real-world articulated objects demonstrate our superiority in generalizable part-level surface reconstruction and joint parameter estimation, compared to existing approaches. Project Site: \href{https://sites.google.com/view/reartgs2/home}{https://sites.google.com/view/reartgs2/home}.

\end{abstract}    
\section{Introduction}
\label{sec:intro}
Articulated objects are ubiquitous in daily life, such as drawers and laptops, and their surface reconstruction holds significant value across various fields, including robotics and embodied intelligence~\cite{wang2025adamanip, Yu2025ArtGS3DGS, yu2024gamma}. Unlike static objects, articulated objects consist of multiple parts with diverse kinematic structures, requiring additional part-level mesh reconstruction and joint parameter estimation. Under this circumstance, ASDF~\cite{ASDF} and DITTO~\cite{jiang2022ditto} train generative models of articulated objects using expensive 3D supervisions. However, these methods exhibit limited generalization to unseen categories due to the variations in shape and motion of articulated objects. To address this limitation, PARIS~\cite{jiayi2023paris} achieves category-agnostic articulated modeling, i.e., part-level reconstruction and motion analysis, using neural implicit radiance fields, only with multi-view RGB images from two different states. Nevertheless, PARIS faces challenges in multi-part perception and smooth dynamic surface reconstruction, suffering from the entangled implicit representation.

\begin{figure}[t]
\centering
\includegraphics[width=1.0\linewidth]{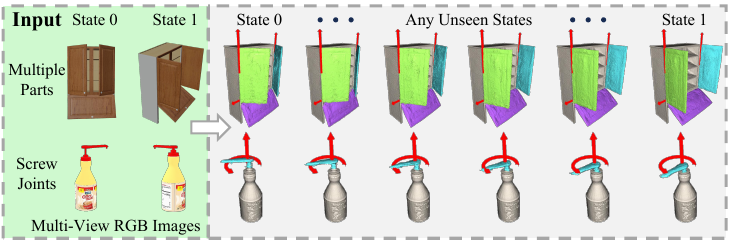}
\caption{Given multi-view RGB images at arbitrary two different states of an unseen articulated objects, our method achieves high-quality part-level dynamic reconstruction and joint parameter estimation, without any external models.}
\vspace{-1.5em}
\label{fig:teaser}
\end{figure}


Recently, ArtGS~\cite{liu2025building} and REArtGS~\cite{wu2025reartgs} achieve efficient and realistic articulated object reconstruction and joint parameter estimation from two-state RGB-D and RGB images respectively, leveraging the explicit representation of 3D Gaussian Splatting (3DGS)~\cite{kerbl3Dgaussians}. However, we observe that these methods still suffer from the following limitations: (1) They rely on joint type priors, which require additional pipelines and pre-defined thresholds to obtain, and the joint types are assumed to be either rotation or translation. It may lead to significant errors when dealing with screw-joint and multi-part objects; (2) It is difficult to impose temporal geometric constraints from only two-state supervisions. Since the joint parameters and Gaussian primitives are jointly optimized, lacking geometric constraint for unseen states futher affects joint parameter estimation. Although REArtGS enhances geometric constraints by incorporating signed distance fields (SDF),  it fails to perform temporal regularization due to the static mapping of SDF.


To address the aforementioned limitations,  we propose REArtGS++, a novel framework of \textbf{R}econstructing g\textbf{E}neralizable \textbf{Art}iculated objects via planar \textbf{G}aussian \textbf{S}platting and temporal geometric constraint, shown as Fig.~\ref{fig:teaser}.  We model a decoupled screw motion for each joint and derive Gaussian's motion from part motion blending, taking part segmentation probability as blending weights. In this way, we jointly optimize part-aware Gaussians and joint parameters without joint type priors. 


To introduce dynamic geometric constraint for improved optimization of articulated modeling, we first compress the 3D Gaussians to flat planes to obtain accurate normals and unbiased depth maps following PGSR~\cite{pgsr}. Then we propose a temporal geometric constraint, extending the consistent regularization of normal and depth from discrete states to entire time interval through Taylor first-order expansion. Moreover, we propose local consistent voting to address the ambiguous segmentation of overlapping regions. Experimental results of both synthetic and real-world data demonstrate our superiority over existing state-of-the-art (SOTA) methods. 

Our main contributions can be summarized as: (1) We model a part-level blending screw motion, achieving unsupervised part segmentation and arbitrary joint parameter estimation, without joint type priors; (2) We propose a temporal geometry constraint for the entire motion process in an unsupervised manner, through Taylor first-order expansion. (3) Extensive experimental results prove that our method is superior to existing SOTA methods in both synthetic and real data.

\section{Related Works}
\label{sec:formatting}
\subsection{Dynamic Reconstruction and Segmentation via 3DGS}
3D Gaussian Spaltting~\cite{kerbl3Dgaussians} and its subsequent works~\cite{pgsr, Huang2DGS2024, yu2024gsdf, scaffoldgs} achieve realistic and efficient scene reconstruction via explicit 3D Gaussian primitives. Recent studies~\cite{yang2023gs4d} have increasingly explored the potential of 3D Gaussian Splatting (3DGS) for dynamic scene reconstruction. Specifically, Deformable-3DGS~\cite{kerbl3Dgaussians} and 4DGS~\cite{4dgs} leverage deformation networks alongside 3D Gaussians, obtaining time-independent Gaussians for dynamic scenes. SC-GS~\cite{huang2023sc} and SP-GS~\cite{spgs} employ superpoints for dynamic Gaussian modeling and control.  Although yielding promising results, each Gaussian learns independent motion in these methods, leading to inconsistency within a rigid part when dealing with articulated objects. Moreover, they require time-continuous supervisions and face challenges in inferring motion using image collections of only two states.

Several works attempt to perform instance segmentation with 3D Gaussians~\cite{zhang2025cob, qin2023langsplat, shi2024language, votesplat, shorinwa2024fast}. These methods achieve 3D Gaussian segmentation by introducing 2D masks obtained from foundation models such as SAM~\cite{kirillov2023segany} or DINO~\cite{zhang2022dino}. However, they rely on the prior knowledge provided by external models and typically struggle with part segmentation. 

\subsection{Articulated Object Reconstruction and Motion Analysis}
Previous methods~\cite{ASDF, song2024reacto} mainly focus on dynamic reconstruction of articulated objects. To achieve both surface reconstruction and motion analysis for articulated objects, PARIS~\cite{jiayi2023paris, weng2024neural} employ composite neural radiance fields to jointly optimize geometry and motion through multi-view RGB images of two different states. Following this setting, several methods~\cite{guo2025articulatedgs, shen2025gaussianart,  liu2025building, wu2025reartgs} introduce 3DGS to articulated object reconstruction and joint parameter estimation. However, these methods require additional pipelines to obtain joint type priors, which are limited to rotation or translation, and affect the performance when dealing with screw-joint or multi-part objects.
\section{Preliminary: Planar Gaussian Spaltting}
3DGS~\cite{kerbl3Dgaussians} uses a set of Gaussians to explicitly represent the scene instead of neural implicit representation. The rendering color can be derived from the $\alpha$-blending as follows:
\begin{equation}
\label{eq:alpha}
	C=\sum_{i =1}^{N_{\mathcal{G}}} \mathbf{c}_{i} \alpha_{i} \prod_{j=1}^{i-1}\left(1-\alpha_{j}\right)
\end{equation} 
where $ N_{\mathcal{G}} $ is the number of ordered points overlapping the pixel, $ \mathbf{c}_{i} $ is the color described by Spherical Harmonics, and $\alpha_{i}$ is obtained from Gaussian's rendering contribution multiplied with a learnable opacity coefficient $\sigma \in\left ( 0,1 \right )  $.  

3DGS learns the 3D Gaussians through a rendering loss $\mathcal{L}_{\text{render}}$, which can be represented as:
\begin{equation}
\label{eq:rendering}
    \mathcal{L}_{\text{render}} = (1-\lambda_{\text{D-SSIM}})\mathcal{L}_{1} + \lambda_{\text{D-SSIM}}\mathcal{L}_{\text{D-SSIM}}
\end{equation}
where $\mathcal{L}_{1}$ is a pixel-level L1 norm and $\mathcal{L}_{\text{D-SSIM}}$ is the D-SSIM loss~\cite{kerbl3Dgaussians}.

Recently, PGSR~\cite{pgsr} obtains accurate normals and unbiased depth estimates by compressing 3D Gaussians to 2D planar representations, enhancing geometric learning. Concretely, the normal maps $\mathbf{N}$ of planar Gaussians can be naturally rendered via their shortest axes and view directions, and the unbiased depth can be estimated as:
\begin{equation}
\mathbf{D}(\mathbf{\rho}) = \frac{\mathbf{d}}{\mathbf{N}(\mathbf{\rho})\mathbf{K}^{-1}{\widetilde{\rho} }}
\label{eq:depth2normal}
\end{equation}
where $\mathbf{d}$ is the distance map from Gaussian's plane to camera center and $\mathbf{K}$ denotes the camera intrinsic matrix. $p$ and $\widetilde{\rho}$ are 2D position and its homogeneous coordinate representation on image plane respectively. The rendered depth is derived from the intersection point of rays and Gaussian flat planes, matching the actual surface without deviation.




\begin{figure*}[ht]
\centering
\includegraphics[width=0.88\linewidth]{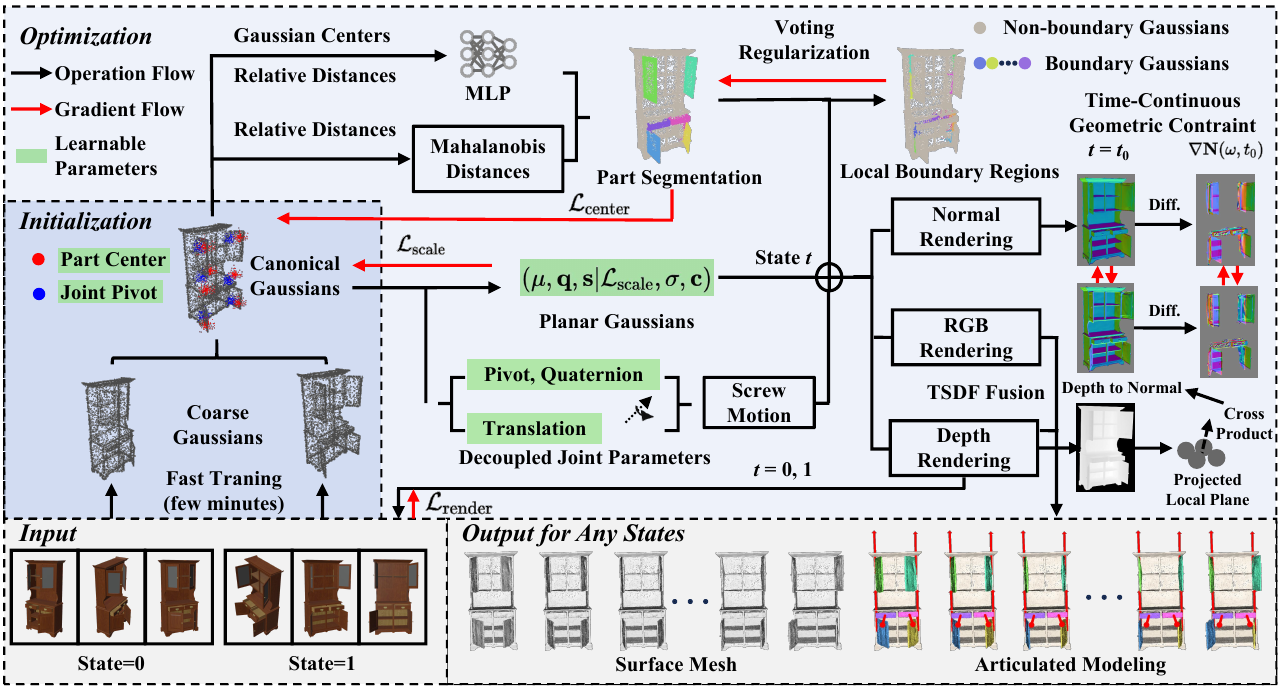}
\caption{Framework of REArtGS++. Our method jointly optimizes part segmentation, joint parameters and planar Gaussians using  multi-view RGB images from arbitrary two states, and achieves high-quality part-level mesh reconstruction of any states and accurate joint parameter estimation for an unseen articulated object. ``Diff.'' denotes the difference approximation.}
\label{fig:method}
\end{figure*} 

\section{Methodology}
Our overall framework is shown in Fig.~\ref{fig:method}. Given an unseen articulated object with $k$ parts, our goal is to achieve part-level surface mesh reconstruction and joint parameter estimation, using multi-view posed RGB  images $\mathbf{I}_{0}, \mathbf{I}_{1}$ in arbitrary two different states without depth supervision. Then we can generate dynamic surface mesh at any state $t=[0, 1]$. The parameters of each joint include rotation angle $\theta \in \mathbb{R}$, rotation axis $\mathbf{a}\in \mathbb{R}^{3}$, pivot point $o\in \mathbb{R}^{3}$, and translation $\mathbf{t}\in \mathbb{R}^{3}$.  Note that the joint may involve any rigid motion (rotation, translation and screw), and we do not need to determine their types before optimization.  The detailed pipelines are elaborated below.

\subsection{Part-Aware Planar Gaussian Representation}
We first introduce a scale loss $\mathcal{L}_{\text{scale}}$ to flatten 3D Gaussian primitives into 2D planes.
\begin{equation}
\mathcal{L}_{\text{scale}} = \frac{1}{N_{\mathcal{G}}} \sum_{i}^{N_{\mathcal{G}}} \left \| \text{min}(s_{1}, s_{2}, s_{3}) \right \| 
\label{eq:scale}
\end{equation}
where $s_{1}, s_{2}, s_{3}$ are the components of Gaussian's scale $\mathbf{s}$ along each direction. In this way, we can obtain accurate normals and unbiased depth maps through Eq.~\ref{eq:depth2normal}, facilitating the temporal geometric constraint in Sec.~\ref{sec:geo}.

Subsequently, we introduce a segmentation mask $\mathbf{M}= \left\{m_{1},m_{2}, ..., m_{k} \right\}$ for each planar Gaussian, representing the probability belonging to each part. Intuitively, the probability that a Gaussian primitive $\mathcal{G}_{i}$ belongs to a part is inversely proportional to its distance from the part's center. Inspired by REACTO~\cite{song2024reacto}, we first denote part centers as learnable parameters $(\mathbf{O}, \mathbf{V},\mathbf{\Lambda} )$, where $\mathbf{O} \in \mathbb{R}^{k \times 3}$, $\mathbf{V} \in \mathbb{R}^{k \times 3\times3}$, $\mathbf{\Lambda} \in \mathbb{R}^{k \times 3}$ are part centers, center orientations and scale vectors respectively. For Gaussian $\mathcal{G}_{i}$, the relative distances $\mathbf{L}_{i}$ between Gaussian center $\mu_{i}$  and part centers are defined as: $\mathbf{L}_{i} = \mathbf{V}(\mu_{i} - \mathbf{O})\mathbf{\Lambda}$. Then the Mahalanobis distances $\mathbf{\gamma}$ can be derived as: $\mathbf{\gamma}_{i} = \mathbf{L}_{i}^{T}\mathbf{L}_{i}$. We calculate part segmentation mask $\mathbf{M}_{i}$ of $\mathcal{G}_{i}$ through the Mahalanobis distances and a residual term learned by an multi-layer perceptron (MLP), which is illustrated in the supplementary materials.



\subsection{Part-level Blending Motion}
First, we briefly analyze the limitations of existing methods in modeling Gaussian motion. REArtGS models Gaussian motion as time interpolation of rotation or translation. ArtGS learns Gaussian motion through dual quaternions, but lacks explicit modeling for the pivot of rotation joints, causing an entangled representation of the pivot and translation. We provide a detailed proof in the supplement. Consequently, these methods struggle with estimating screw motion and cause additional errors when pre-determining joint types for multi-part objects.

To model a screw motion SE (3) without prior, we decouple the parameters $\omega$ of each joint into rotation and translation and set $\omega = \left\{\mathbf{q}\left(\theta, \mathbf{a}\right), \mathbf{o}, \mathbf{t}\right\} $ as trainable parameters for each part. For simplicity, we use $\mathbf{q}$ to represent joint quaternion $\mathbf{q}\left(\theta, \mathbf{a}\right)$ in the following. Then the motion of planar Gaussian $\mathcal{G}_{i}$ can be derived from the blending of part motions, where the segmentation probability $\left\{m_{1}^{i}, m_{2}^{i}, ..., m_{k}^{i}\right\}$ serves as the blending weights. We formulate the transformed position $\mu_{i}(t)$ as following:
\begin{equation}
\label{eq:motion}
    \mu_{i}(t) = \sum_{j=1}^{k} m_{j}    \left [   R_{j}(\mathbf{q}(t))(\mathbf{\mu}_{i}-\mathbf{o}_{j})+\mathbf{o}_{j}+\mathbf{t}_{j}(t)   \right ] 
\end{equation}
where $t\in[0,1]$ denotes the state, and $R(\mathbf{q}(t))$ and $\mathbf{t}(t)$ is the time-varying rotation matrix and translation. We illustrate their calculation in detail below.

We define the canonical state as $t^{*}=0.5$ to ensure the rotation angle $\theta$  belongs to $\left [ - \pi /2, \pi /2 \right ] $, preventing the singularity in exponential coordinates when $\left \| \theta  \right \| > \pi$, following REArtGS. The rotation angle and translation at state $t$ can be represented as:
\begin{equation}
\theta(t) =  \frac{\left ( t -t^{*} \right ) }{t^{*}} \theta, \quad\mathbf{t} = \frac{(t-t^{*})}{t^{*}} \mathbf{t}
\end{equation}
and the time-varying rotation quaternion can be formulated as: 
\begin{equation}
 \mathbf{q}(t) = \cos(\frac{\theta(t)}{2}) + \mathbf{a}\sin(\frac{\theta(t)}{2})
\end{equation}
Hence, the rotation matrix $R(\mathbf{q}(t))$ can be derived through $\mathbf{q}(t)$. In this way, we jointly optimize the part segmentation $\mathbf{M}$ and joint parameters in a unsupervised manner.

\begin{table*}[t!]
\centering
\caption{\textbf{Quantitative results on PARIS dataset.} We implement all methods without depth supervision for fair comparison. Axis Pos results are measured by mm. "-" indicates the object containing only prismatic joints. We highlight \colorbox[HTML]{ffc5c5}{best} and \colorbox[HTML]{ffebd8}{second best} results.}
\label{tab:exp_paris}
\resizebox{\linewidth}{!}{
\begin{tabular}{cc|ccccccccccc|ccc}
\hline

\multirow{2}{*}{Metric} &\multirow{2}{*}{Method} &\multicolumn{11}{c}{Synthetic Objects} &\multicolumn{3}{|c}{Real Objects} \\
& &FoldChair &Fridge &Laptop &Oven &Scissor &Stapler &USB 
&Washer&Blade &Storage &Mean & Fridge &Storage &Mean \\ 
\hline
\multirow{5}{*}{\shortstack{Axis\\Ang}} 
&PARIS~\cite{jiayi2023paris} &15.72 &2.02 &0.03 &7.43 &16.62 &7.17 &0.71 &15.40 &38.28 &0.03 &10.34 &2.90 &20.16 &11.53\\
&DTA~\cite{weng2024neural} &0.03 &0.08 &0.07 &0.21 &0.10 &0.08 &0.11 &0.38 &0.18 &0.09 & 0.13 &2.26 &10.57 &6.42\\
&ArtGS~\cite{liu2025building} &\secbest{0.01} &0.03 &\best{$<$0.01} &\secbest{0.02} &\best{$<$0.01} &\secbest{0.01} &\best{0.02} &\secbest{0.04} &\secbest{0.02} &\secbest{0.01} &\secbest{0.02} &\secbest{2.03} &\secbest{4.56} &\secbest{3.30}\\
&REArtGS~\cite{wu2025reartgs} &\secbest{0.01} &\secbest{0.02} &0.02 &0.07 &0.05 &0.06 &0.04 &0.08 &0.04 &0.04 &0.04 &2.19 &87.51 & 44.85\\
&REArtGS++ (Ours)
&\best{$<$0.01} &\best{0.01} &\best{$<$0.01} &\best{$<$0.01} &\secbest{0.04} &\best{$<$0.01} &\secbest{0.04} &\best{0.02} &\best{0.01} &\best{$<$0.01} &\best{0.01} &\best{1.92} &\best{3.34} &\best{2.63}\\
\hline
\multirow{5}{*}{\shortstack{Axis\\Pos}} 
&PARIS~\cite{jiayi2023paris} &10.30 &9.18 &\secbest{0.76} &5.15 &15.31 &35.07 &13.26 &24.14 &- &- &14.15 &50.28 &- &50.28\\
&DTA~\cite{weng2024neural} &1.47 &1.16 &2.32 &\secbest{1.18} &1.95 &1.67 &1.46 &4.87 &- &- &2.01 &59.13 &- &59.13\\
&ArtGS~\cite{liu2025building} &\best{0.06} &0.43 &0.94 &\secbest{1.35} &\secbest{0.18} &\best{0.38} &\secbest{0.15} &1.06 &- &- &\secbest{0.51} &\secbest{42.83} &- &\secbest{42.83}\\
&REArtGS~\cite{wu2025reartgs} &6.24 &\best{0.15} &1.31 &3.64 &0.06 &0.72 &1.56 &\best{0.04} &- &- &1.72 &45.72 &- &45.72\\
&REArtGS++ (Ours) &\secbest{0.37} &\secbest{0.34} &\best{0.73} &\best{0.71} &\best{$<$0.01} &\secbest{0.49} &\best{0.04} &\secbest{0.95} &- &- &\best{0.45} &\best{42.66} &- &\best{42.66}\\
\hline
\multirow{5}{*}{\shortstack{Part\\Motion}}
&PARIS~\cite{jiayi2023paris} &45.80 &68.21 &0.03 &9.70 &71.42 &128.73 &34.57 &43.58 &0.45 &0.30 &40.28 &\secbest{2.23} &0.57 &1.40\\
&DTA~\cite{weng2024neural} &0.10 &0.12 &0.11 &0.12 &0.34 &0.08 &0.15 &0.24 &\best{$<$0.01} &\best{$<$0.01} &0.13 &2.56 &0.18 &1.37\\
&ArtGS~\cite{liu2025building} &\secbest{0.03} &\secbest{0.03} &\secbest{0.03} &\secbest{0.01} &\best{0.01} &\secbest{0.03} &\best{$<$0.01} &\secbest{0.02} &\best{$<$0.01} &\secbest{0.01} &\secbest{0.02} &2.27 &\secbest{0.15} &\secbest{1.21}\\
&REArtGS~\cite{wu2025reartgs} &0.31 &0.23 &0.06 &0.08 &0.06 &0.15 &0.05 &0.04 &\best{$<$0.01} &0.08 &0.11 &38.38 &0.52 &19.45\\
&REArtGS++ (Ours) &\best{0.02} &\best{$<$0.01} &\best{0.02} &\best{$<$0.01} &\secbest{0.03} &\best{0.01} &\secbest{0.03} &\best{0.01} &\best{$<$0.01} &\best{$<$0.01} &\best{0.01} &\best{2.16} &\best{0.13} &\best{1.15}\\
\hline
\multirow{5}{*}{\shortstack{CD-s}} 
&PARIS~\cite{jiayi2023paris} &8.51 &6.30 &0.25 &\secbest{4.57} &12.71 &2.85 &1.95 &\best{8.37} &10.64 &8.62 &6.48 &10.74 &27.61 &19.18\\
&DTA~\cite{weng2024neural} &\secbest{0.27} &2.48 &0.64 &10.06 &1.64 &3.08 &2.64 &20.73 &\secbest{0.46} & 10.73 &5.74 &4.50 &30.98 &17.74\\
&ArtGS~\cite{liu2025building} &\best{0.16} &1.89 &0.80 &8.58 &0.50 &\secbest{3.55} &\secbest{2.51} &23.63 &0.51 &11.82 &5.40 &\secbest{3.39} &\secbest{18.19} &\secbest{10.79}\\
&REArtGS~\cite{wu2025reartgs} &0.31 &\secbest{1.53} &\best{0.48} &6.59 &\best{0.25} &\best{1.47} &\best{0.82} &16.04 &0.69 &\best{7.47} &\secbest{3.57} &7.67 &22.53 &15.10\\
&REArtGS++ (Ours)  &\secbest{0.27} &\best{0.97} &\secbest{0.53} &\best{4.14} &\secbest{0.37} &3.60 &3.12 &\secbest{10.59} &\best{0.39} &\secbest{10.41} &\best{3.44} &\best{1.33} &\best{10.98} &\best{6.16}\\ 
\hline
\multirow{5}{*}{\shortstack{CD-m}}
&PARIS~\cite{jiayi2023paris} &4.05 &1.26 &\best{0.15} &3.70 &10.06 &1.18 &3.02 &1.52 &5.13 &20.67 &5.07 &74.20 &314.37 &194.29\\
&DTA~\cite{weng2024neural} &\secbest{0.46} &2.27 &0.96 &1.36 &10.25 &1.61 &1.37 &4.45 &2.07 &\secbest{5.34} &3.01 &3.63 &81.36 &42.50\\
&ArtGS~\cite{liu2025building} &0.52 &1.40 &0.98 &2.52 &0.70 &\secbest{0.86} &\secbest{1.15} &4.67 &\best{1.23} &6.56 &2.06 &3.23 &\secbest{33.65} &\secbest{18.44}\\
&REArtGS~\cite{wu2025reartgs} &\best{0.30} &\best{0.66} &0.71 &\secbest{1.25} &\best{0.25} &\best{0.21} &\best{0.40} &\best{0.42} &\secbest{1.97} &12.39 &\secbest{1.85} &\secbest{2.87} &175.78 &89.33\\
&REArtGS++ (Ours) &0.77 &\secbest{1.25} &\secbest{0.60} &\best{0.85} &\secbest{0.83} &0.90  &\secbest{1.15} &\secbest{0.89} &2.54 &\best{4.78} &\best{1.46} &\best{1.75} &\best{13.85} &\best{7.80}\\

\hline
\multirow{5}{*}{\shortstack{CD-w}}
&PARIS~\cite{jiayi2023paris} &8.27 &2.83 &3.13 &11.08 &11.68 &2.80 &12.46 &19.67 &0.81 &8.37 &8.11 &8.82 &24.73 &16.78\\
&DTA~\cite{weng2024neural} &0.58 &2.36 &0.75 &9.35 &0.65 &2.95 &\secbest{1.43} &18.68 &0.51 &8.34 &4.56 &4.47 &23.48 &13.98\\
&ArtGS~\cite{liu2025building} &\best{0.44} &2.01 &0.96 &\secbest{8.32} &0.72 &\best{2.71} &1.57 &21.69 &0.67 &8.69 &4.78 &3.26 &12.26 &15.52\\
&REArtGS~\cite{wu2025reartgs} &\secbest{0.70} &\secbest{1.64} &\secbest{0.65} &10.89 &\best{0.31} &3.26 &1.54 &\secbest{17.82} &\best{0.40} &\secbest{7.71} &\secbest{4.49} &\secbest{2.93} &\secbest{11.48} &\secbest{7.21}\\
&REArtGS++(Ours)  &0.51 &\best{1.20} &\best{0.48} &\best{3.72} &\secbest{0.56} &\secbest{2.87} &\best{1.30} &\best{13.97} &\secbest{0.50} &\best{7.39} &\best{3.25} &\best{1.52} &\best{10.63} &\best{6.08}\\

\hline
\end{tabular}
}
\vspace{-10pt}
\end{table*}

\begin{figure*}[tp]
\centering
\includegraphics[width=0.88\linewidth]{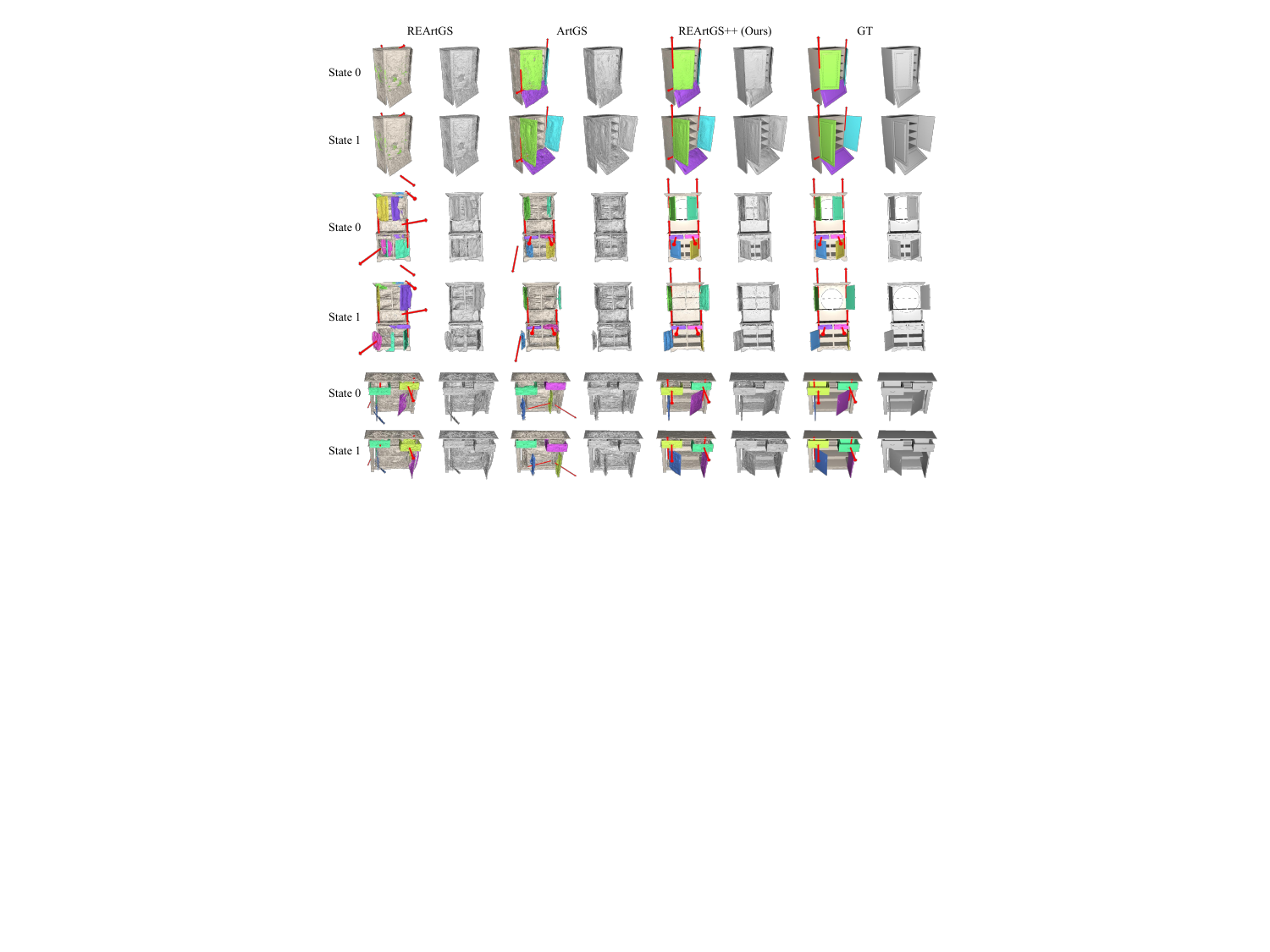}
\caption{The qualitative results of dynamic surface reconstruction at start state and end state on ArtGS-Multi dataset. We show both part segmentation and surface meshes for best comparison. The red arrows represent joints.}
\label{fig:artgs_multi1}
\vspace{-1.0em}
\end{figure*} 


\subsection{Temporal Geometric Constraint}
\label{sec:geo}
To enhance the geometric constraint in dynamic rendering, we propose an unsupervised consistency regularization between depth and normals of Gaussians. Leveraging the joint parameters $\omega$, we can obtain the normal map $\mathbf{N}(\omega, t)$  through $\alpha$-blending similar to Eq.~\ref{eq:alpha}, and depth map $\mathbf{D}(\omega, t)$ via Eq.~\ref{eq:depth2normal} at any state $t\in[0,1]$. Inspired by PGSR, we take the Gaussian as a local plane and sample a local patch (4 neighboring pixels) from $\mathbf{D}(\omega, t)$ and project them into 3D points. The estimated normal can be derived from their cross product, and we acquire estimated normal map $\bar{\mathbf{N}}(\omega,t)$ by repeating the process for entire depth map. A natural technique is to align $\bar{\mathbf{N}}$ with $\mathbf{N}$, as in PGSR. However, this approach is limited to a single discrete state and cannot regularize the entire interval $t \in [0,1]$. To solve the above problem, we adopt the Taylor first-order expansion to approximate $\mathbf{N}(\omega, t)$, which can be formulated as:
\begin{equation}
    \mathbf{N}(\omega, t) \approx  \mathbf{N}(\omega, t_{0})+\lim_{t \to t_{0}}\frac{\mathrm{d} N(\omega ,t)}{\mathrm{d} t}  (t-t_{0})
\end{equation}
where $t_{0} = 0, 1$ to ensure that the depth and normal renderings can be obtained simultaneously through a single forward when calculating color rendering loss. Furthermore, we use a difference approximation to represent the gradient of the normal against time, shown as:
\begin{equation}
\label{eq:diff}
  \nabla\mathbf{N}(\omega, t_{0}) =\lim_{t \to t_{0}}\frac{\mathrm{d} N(\omega ,t)}{\mathrm{d} t} \approx  \frac{N(\omega ,t)-N(\omega ,t^{*})}{t-t^{*}}
\end{equation}
Since the Gaussians do not apply motion transformation at canonical state $t^{*}$, this approximation reduces computational costs and prevents the accumulation errors of motion during optimization. Note that $N(\omega ,t^{*})$ does not participate in back-propagation to save memory.
The same approximation is adopted for $\bar{\mathbf{N}}(t)$. In this way, the time-continuous geometry constraint for the entire interval $t \in [0,1]$ can be represented as following:
\begin{equation}
\begin{aligned}
\label{eq:geo}
\mathcal{L}_{geo} &= (1-\nabla \mathbf{I}(t_{0}))\left( \left\| \bar{\mathbf{N}}(\omega, t_{0})-\mathbf{N}(\omega, t_{0}) \right\| \right. \\ 
              &\left. +\left\| \nabla\bar{\mathbf{N}}(\omega, t_{0})-\nabla\mathbf{N}(\omega, t_{0}) \right\| \right) 
\end{aligned}
\end{equation}
where $\nabla \mathbf{I}(t)$ is the image gradient to avoid computations at the edges of images.

\subsection{Local Consistent Voting}

\label{sec:local_consistency_voting}
We observe the distance-based segmentation $\mathbf{M}$ often fails to learn a probability distribution with clear tendencies at boundary Gaussians $\hat{\mathcal{G}}$ between adjacent parts, leading to ambiguous and inconsistent part segmentation in such regions. To address this, we introduce a local consistent voting mechanism to constrain the consistency of part segmentation for boundary Gaussians $\hat{\mathcal{G}}$.

 
Given a part $j$, we define a Gaussian $\mathcal{G}_{i}$ within it as a boundary Gaussian when meeting the following condition:
\begin{equation}
\sum\frac{\Gamma \left(\text{max}(\mathbf{M}_{\Omega (\mathcal{G}_{i})}) \ne  m_{j}\right)}{\left | \Omega (\mathcal{G}_{i}) \right | } \ge \beta
\end{equation}
where $\Omega (\mathcal{G}_{i})$ is the neighbor Gaussians of $\mathcal{G}_{i}$ sampled by K-Nearest Neighbor (KNN) and $m_{j}$ denotes the probability belonging to part $j$. $\Gamma$ is an indicator function and $\beta$ is set to 0.2.

The boundary Gaussians are then divided into $4\times k$ local regions by k-means algorithm. For each region, we derive a voting segmentation probability $\mathbf{M}_{\text{vote}}$ through calculating the probability distributions $\mathbf{M}$ of all points, weighted by their distance to regional center. We formulate the calculation as following:
\begin{equation}
    \mathbf{M}_{\text{vote}} = \sum_{i}^{\mathcal{N}(\mathcal{G}_i)} \frac{\Phi_{\text{softmax}}(-\delta  ) M_{i}}{\sum_{i}^{\mathcal{N}(\mathcal{G}_i)} M_{i}} 
\end{equation}
where $\Phi_{\text{softmax}}$ is the Softmax activation and $\delta$ represents the distance from Gaussian $\mathcal{G}_{i}$ to regional center.

Finally, we constrain the local segmentation consistency by minimizing the divergence between learned distribution for a boundary Gaussian and the voting distribution, which can be represented as:
\begin{equation}
\mathcal{L}_{\text{vote}} = \sum_{n}^{|\mathcal{N}|} \sum_{\mathcal{G}_i \in \mathcal{N}_{n }(\mathcal{G}_i)}\frac{1}{|\mathcal{N}_{n}(\mathcal{G}_i)|} D_{\text{KL}}\left( M_{\text{vote}} \,\|\, M_i \right)
\label{eq:vote}
\end{equation}
where $D_{\text{KL}}(\cdot\|\cdot)$ denotes the KL divergence and $|\mathcal{N}|$ is the number of local regions. Intuitively, the regularization introduces spatial context information for part segmentation by aggregating probability distributions of local regions, instead of relying solely on point-wise understanding.


\subsection{Initialization and Optimization}
\label{sec:Initialization and Optimization}
Inspired by ArtGS, we first perform a fast optimization for two sets of Gaussians at both start and end states using vanilla 3DGS pipeline~\cite{kerbl3Dgaussians}, then identify the dynamic Gaussians for each state by criterion: $\Delta \textbf{x}_{i} > \tau\textbf{x}$, where $\Delta \textbf{x}_{i}$ is the Chamfer Distance (CD) from Gaussian $\mathcal{G}_{i}$ at one state to all Gaussians of another state. $\textbf{x}$ is the maximum CD and $\tau$ is set to 0.02. We initialize static and dynamic Gaussians of canonical state as the mean of the two states, using Hungarian matching.

Taking the static Gaussian as a cluster, we adopt K-means clustering algorithm to divide dynamic Gaussians into $k-1$ clusters and set cluster centers as the initialization of $k$-part centers. We set the initial scale of each part as the maximum distance from the farthest point to the cluster center. Given a cluster of part ${j}$, we extract the Gaussians of the connected regions between the static part and part ${j}$ through a similar mechanism in Sec.~\ref{sec:local_consistency_voting}. Then the mean position of the Gaussians within a connected region serves as initial pivot point for corresponding joint.

Once the initialization is completed, we optimize the dynamic rendering of planar Gaussians through the multi-view images at two states. The rendering loss $\mathcal{L}_{\text{render}}$ can be calculated through Eq.~\ref{eq:rendering} using $\mathbf{I}_{t}$ and $\bar{\mathbf{I}}(\omega, t)$, where $\bar{\mathbf{I}}$ denotes the rendered images and $t = 0,1$. In this way, our method also supports multi-state observation inputs $\left\{\textbf{I}_{0},\textbf{I}_{t_{1}}, \textbf{I}_{t_{2}}, ...,\textbf{I}_{1}\right\}$ by extending $t$ to $\left\{0, t_{1}, t_{2}, ..., 1\right\}$. Please refer to the supplement for more details. Following ArtGS, we also introduce a center regularization $\mathcal{L}_{\text{center}}$ to align part centers with the mean positions of the corresponding Gaussians.

In summary, our total training objective can be represented as following:
\begin{equation}
\begin{aligned}
    \mathcal{L} &= \lambda_{\text{render}} \mathcal{L}_{\text{render}} + 
    \lambda_{\text{scale}} \mathcal{L}_{\text{scale}} +
    \lambda_{\text{center}} \mathcal{L}_{\text{center}} \\
    &+  \lambda_{\text{geo}} \mathcal{L}_{\text{geo}} + \lambda_{\text{vote}} \mathcal{L}_{\text{vote}}
\end{aligned}
\end{equation}

\subsection{Dynamic Mesh Extraction}
Once the optimization converges, we can extract dynamic surface meshes at any state $t \in [0,1]$ via the planar Gaussians $\mathcal{G}$ and joint parameters $\left \{\omega_{1},...,\omega_{k}\right\}$. Specifically, we obtain the Gaussians $\mathcal{G}^{j}$ of a given part $j$ by: 
\begin{equation}
   \mathcal{G}^{j}=\left\{ \mathcal{G}_{i}|\text{max}(\mathcal{G}_{i}=m_{j}), i\in N_{\mathcal{G}} \right\} 
\end{equation}
Then the dynamic Gaussians $\mathcal{G}^{j}$ at state $t$ can be updated by Eq.~\ref{eq:motion}.
The RGB and depth renderings of each part can be directly obtained by Eq.~\ref{eq:alpha} and Eq.~\ref{eq:depth2normal}. We employ the TSDF fusion algorithm to extract part-level dynamic meshes using these maps as inputs, where the voxel size is set to 0.04.

\begin{table*}[htbp]
\centering
\small

\caption{\textbf{Quantitative results on ArtGS-Multi dataset with additional 2 objects exhibiting screw joints}. Due to the large number of parts, we report the average metric for all movable parts. We implement all methods without depth supervision for fair comparison. Part motion error for screw objects is decomposed by translation$|$rotation. Axis Pos results are measured by mm. `-' indicates the object contains only prismatic joints. We highlight \colorbox[HTML]{ffc5c5}{best} and \colorbox[HTML]{ffebd8}{second best} results.}
\label{tab:exp_mpart}

\resizebox{0.95\linewidth}{!}{
\begin{tabular}{cccccccc|ccc}
\toprule
\multirow{2}{*}{Metrics} & \multirow{2}{*}{Method}
&Table &Table &Storage &Storage &Oven & \multirow{2}{*}{Mean} &Bottle &Dispenser &\multirow{2}{*}{Mean} \\
& &\scriptsize{25493 (4 parts)} &\scriptsize{31249 (5 parts)} &\scriptsize{45503 (4 parts)} &\scriptsize{47468 (7 parts)} &\scriptsize{101908 (4 parts)} & &\scriptsize{3398 (screw part)} &\scriptsize{103378 (screw part)} & \\

\midrule
\multirow{3}{*}{\shortstack{Axis\\Ang}}
&ArtGS~\cite{liu2025building} &\secbest{1.23} &83.69 &\best{0.01} &34.83 &\secbest{5.62} &\secbest{25.08} &\secbest{42.28} &\secbest{87.57} & \secbest{64.93}\\
&REArtGS~\cite{wu2025reartgs}   &85.45 &\secbest{42.18} &64.34 &30.69 &24.50 &49.43 &46.45 &87.80 &67.13\\
&REArtGS++ (Ours)  &\best{0.02} &\best{0.05} &\secbest{0.06} &\best{2.58} &\best{0.03} &\best{0.55} &\best{0.01} &\best{0.12} &\best{0.07}\\
\midrule
\multirow{3}{*}{\shortstack{Axis\\Pos}}
&ArtGS~\cite{liu2025building} & - &\secbest{70.93} &\secbest{0.36} &235.19 &8.54 &\secbest{96.49} &50.64 &38.75 &44.70\\
&REArtGS~\cite{wu2025reartgs} &- &155.80 &500.51 &\secbest{31.44} &233.48 &230.31 &\secbest{47.18} &\secbest{20.41} &\secbest{33.80}\\
&REArtGS++ (Ours) &- &\best{0.10} &\best{0.21} &\best{0.06} &\best{1.62} &\best{0.50} &\best{1.04} &\best{4.03} &\best{2.54}\\
\midrule
\multirow{3}{*}{\shortstack{Part\\Motion}}
&ArtGS~\cite{liu2025building} &\best{0.08} &22.50 &\best{0.01} &\secbest{18.75} &\secbest{5.73} &\secbest{9.41} &0.06$|$65.87 &\secbest{0.02$|$31.84} &\secbest{0.04$|$48.86}\\
&REArtGS~\cite{wu2025reartgs}  &\secbest{0.23} &\secbest{13.09} &46.09 &25.41 &55.53 &28.07 &\secbest{0.04$|$63.74} &0.03$|$45.90 &\secbest{0.04}$|$54.82\\
&REArtGS++ (Ours)  &0.67 &\best{0.09} &\secbest{0.08} &\best{0.05} &\best{0.21} &\best{0.22} &\best{0.01$|$0.02} &\best{0.01$|$0.44} &\best{0.01$|$0.23}\\
\midrule
\multirow{3}{*}{CD-s}
&ArtGS~\cite{liu2025building} &85.65 &3.01 &\secbest{2.35} &\secbest{1.74} &\secbest{10.45} &20.64 &1.78 &0.30 &1.04\\
&REArtGS~\cite{wu2025reartgs} &\secbest{45.11} &\secbest{2.86} &14.12 &2.16 &15.83 &\secbest{16.02} &\secbest{0.85} &\secbest{0.21} &\secbest{0.53}\\
&REArtGS++ (Ours) &\best{0.77} &\best{1.26} &\best{1.35} &\best{0.61} &\best{5.03} &\best{1.80} &\best{0.64} &\best{0.18} &\best{0.41}\\
\midrule
\multirow{3}{*}{CD-m}
&ArtGS~\cite{liu2025building} & 216.58 &\secbest{9.50} &\best{0.39} &\secbest{7.49} &\best{0.79} &\secbest{46.95} &\secbest{0.46} &11.75 &\secbest{6.11} \\ 
&REArtGS~\cite{wu2025reartgs} & \secbest{71.44} &10.63 &119.32 &70.91 &136.68 &81.80 &0.97 &\secbest{10.74} &40.86 \\
&REArtGS++ (Ours) & \best{1.39} &\best{4.89} &\secbest{0.58} &\best{3.19} &\secbest{0.80} &\best{2.17} &\best{0.08} &\best{5.70} &\best{2.89}\\
\midrule
\multirow{3}{*}{CD-w}
&ArtGS~\cite{liu2025building} &\secbest{1.06} &\secbest{4.06} &\best{2.29} &\secbest{3.31} &\secbest{9.87} &\secbest{4.12} &1.75 &0.59 &1.17\\
&REArtGS~\cite{wu2025reartgs} &3.52 &2.88 &13.16 &6.16 &17.96 & 8.74&\secbest{0.80} &\secbest{0.47} &\secbest{0.64}\\
&REArtGS++ (Ours) &\best{0.72} &\best{1.13} &\secbest{3.76} &\best{2.13} &\best{5.14} &\best{2.58} &\best{0.59} &\best{0.31} &\best{0.45}\\
\bottomrule 
\end{tabular}
}
\vspace{-5pt}
\end{table*}

\section{Experiment}
\subsection{Experimental Setting}

\textbf{Datasets.} We compare our method with existing SOTAs on three datasets: (1) PARIS dataset~\cite{jiayi2023paris}, including 10 objects from PartNet-Mobility dataset~\cite{SAPIEN}; (2) ArtGS-Multi~\cite{liu2025building}, including 5 objects from PartNet-Mobility dataset, and We collect 2 additional screw joint objects from PartNet-Mobility dataset; (3) Real-world dataset, containing 2 objects from PARIS and 3 objects collected by ours. Each state provides 60-100 posed RGB images. 

\textbf{Metrics.} Following ArtGS, we generate meshes of both start and end states from canonical state, and report the mean Chamfer Distance (CD) resuls, which are multiplied by 1,000. We employ CD-w, CD-s, and CD-m for whole surfaces, static parts and dynamic parts respectively. We evaluate the predicted joints using angular error (Axis Ang.) and pivot distance (Axis Pos., except for translation joints), which are measured by mm, i.e., multiplied by 100. We also report part motion error, which quantifies the geodesic rotation error  (in degrees) and the Euclidean translation distance error (in meters).  Lower ($\downarrow$) is better for all metrics.

All experiments are conducted on a single RTX 4090 GPU. Both our method and ArtGS are optimized for 30,000 iterations, while REArtGS requires an additional 30,000 iterations for its first stage. Please refer to the supplement for detailed implementation.

\begin{figure}[htp]
\centering
\includegraphics[width=1\linewidth]{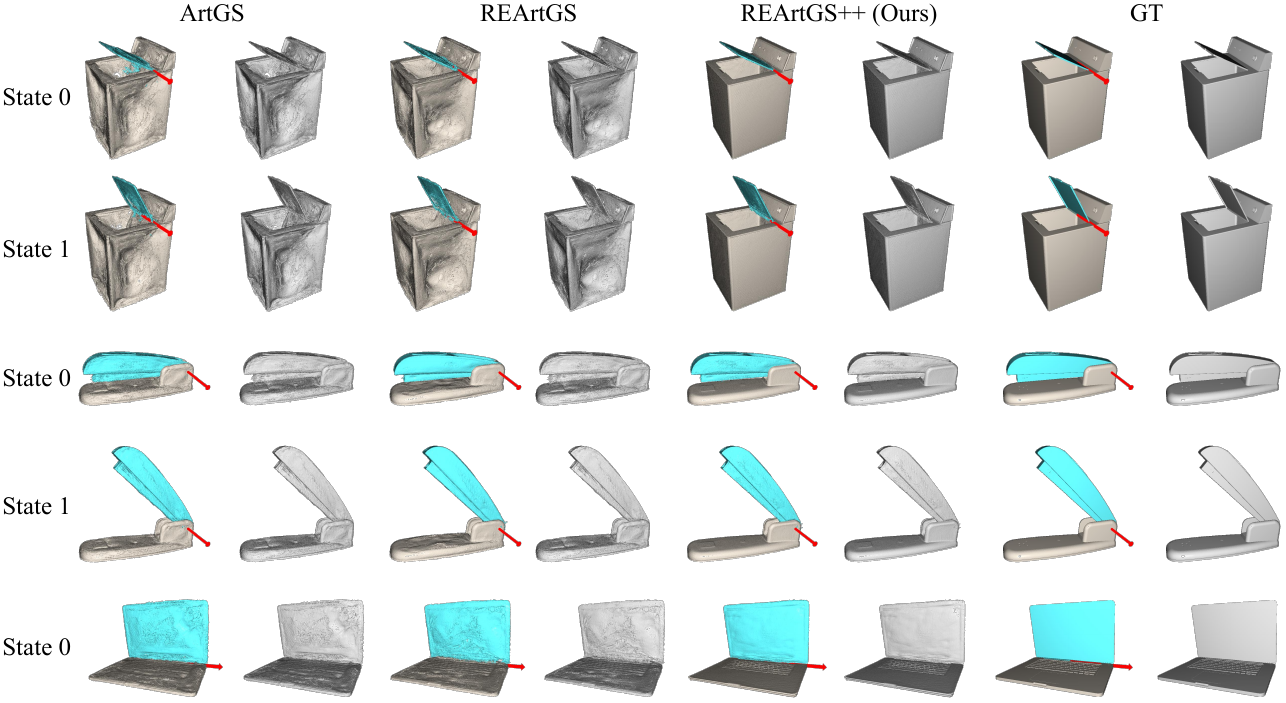}
\caption{The qualitative results of dynamic surface reconstruction at start state and end state on PARIS dataset. We show both articulated modeling and surface meshes for best comparison.}
\label{fig:paris1}
\vspace{-1em}
\end{figure} 

\begin{figure}[h]
\centering
\includegraphics[width=1\linewidth]{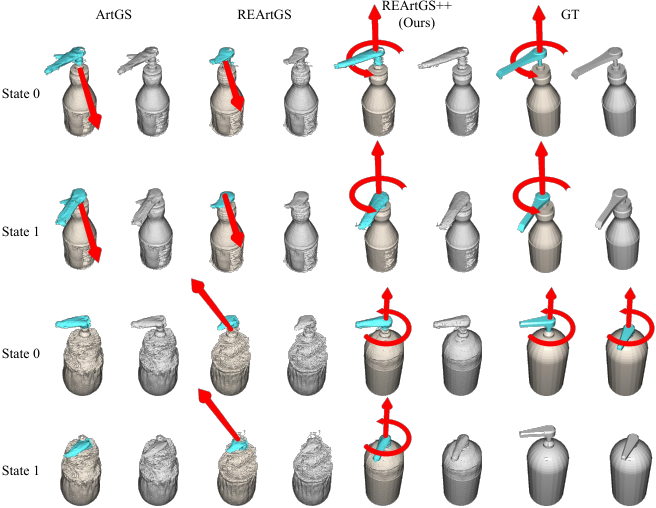}
\caption{The qualitative results of dynamic surface reconstruction at start state and end state on PARIS dataset. We show both articulated modeling and surface meshes for best comparison.}
\label{fig:screw}
\vspace{-2em}
\end{figure}

\subsection{Experiments on Two-Part Articulated Objects}
We present the quantitative and qualitative results on the PARIS dataset in Table.~\ref{tab:exp_paris} and Fig.~\ref{fig:paris1} respectively. More qualitative results can be found in the supplementary material. Our method achieves the best results on the mean values of all metrics especially in CD-w of Synthetic objects, achieving a \textbf{27.6}\% error reduction compared to REArtGS. This is primarily attributed to the enhancement in dynamic reconstruction quality through the dynamic geometric constraints.  In addition, The superior performance of the joint parameter estimation results demonstrates the effectiveness of decoupling screw motion learning.

\subsection{Experiments on Screw and Multi-Part Articulated Objects}
The quantitative and qualitative results of screw and multi-part objects are provided in Table.~\ref{tab:exp_mpart}. We also provide the qualitative results in Fig.~\ref{fig:artgs_multi1} and Fig.~\ref{fig:screw} respectively. Our method exhibits significant advantages over existing methods in the mean values of all metrics, especially in Axis Ang, Axis Pos and Part Motion, achieving mean results of \textbf{0.55}, \textbf{0.50}, \textbf{0.22} respectively.  ArtGS and REArtGS rely on prior knowledge of joint types, where incorrect priors severely degrade subsequent joint axis parameter estimation (e.g., \textit{Table 31249} and \textit{Storage 47468}), while our method does not require joint type estimation, exhibiting greater robustness for multi-part objects.


\textbf{Optimization time analysis}.
On a single RTX 4090 GPU, our method is comparable to ArtGS in terms of computational efficiency, taking an average of 16 to 20 minutes for each object approximately, which is significantly faster than REArtGS (about 70 minutes).

\begin{figure}[h]
\centering
\includegraphics[width=0.8\linewidth]{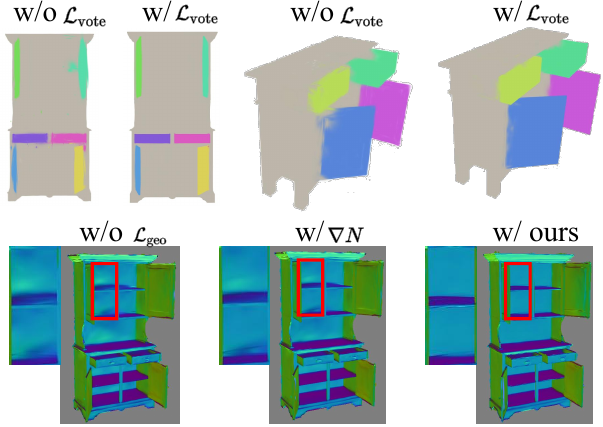}
\caption{Qualitative results for the ablation of voting and geometric regularization. We use segmentation rendering results for intuitive visualization.}
\label{fig:ablation}
\vspace{-1.0em}
\end{figure} 

\begin{table}[h]
\centering
\small

\caption{\textbf{Ablation of key componets}. We report the average results on ArtGS-Multi and 2 screw-joint objects. Axis Pos results are measured by mm. ``GS" and ``dis" denote Gaussians and distance respectively.}
\label{tab:ablation_kc}

\resizebox{\linewidth}{!}{
\begin{tabular}{ccccccccc}
\toprule
Setting 
&Axis Ang &Axis Pos &Part Motion &CD-s &CD-m &CD-w \\
\midrule
w/o planar GS & 5.18 &21.38 &6.23 &5.84 &7.57 &6.45 \\
w/o screw motion &22.78 &88.26 &18.04 &16.24 &29.68 &3.16 \\
w/o MLP &20.72 &34.01 &27.52 &14.38 &76.50 &2.72\\
w/o relative dis. &57.81 &38.17 &35.17 &7.34 &93.80 &6.38 \\
w/o initialization &33.47 &71.74 &40.13 &6.81 &126.03 &8.43 \\
w/o $\mathcal{L}_{\text{center}}$ &28.48 &47.36 &12.63 &3.65 &31.54 &3.32\\
w/o $\mathcal{L}_{\text{vote}}$ & 2.66 &15.10 &5.36 &2.71 &10.38 &3.07 \\
w/ all & \best{0.41} &\best{1.18} &\best{0.22} &\best{1.41} &\best{2.38} &\best{2.13}\\
\bottomrule 
\end{tabular}
}
\vspace{-5pt}
\end{table}

\subsection{Ablation Study}
\textbf{Ablation of key components}. To further validate the effectiveness of our key components, we conducted ablation studies on ArtGS-Multi dataset with 2 additional objects exhibiting screw joints. We present the quantitative results in Table.~\ref{tab:ablation_kc} and a qualitative comparison of local voting regularization in Fig.~\ref{fig:ablation}. For ``w/o planar GS", we employ the original 3DGS pipeline~\cite{kerbl3Dgaussians}; for ``w/o screw motion", joint motions are represented with dual quaternions; for ``w/o initialization", all part centers and joint pivots are randomly initialized. Experimental results demonstrate that each component contributes significantly to overall performance. 

\textbf{Ablation of temporal geometry constraints}. To validate the effectiveness of the temporal geometric constraints, we conduct ablation studies on ArtGS-Multi dataset with 2 additional objects exhibiting screw joints. We present the quantitative results in Table.~\ref{tab:ablation_geo} and a qualitative comparison in Fig.~\ref{fig:ablation}. For ``w/o  $\nabla N$", $\nabla N$ and $\nabla \bar{N}$ are ignored in Eq.~\ref{eq:geo}; For ``w/ random $\Delta t$" and ``w/ tiny $\Delta t$", we use $t_{0} \pm 0.1$ and a random $t_{\text{rand}} \in [0,1]$ to replace $t^{*}$ in Eq.~\ref{eq:diff}. Experimental results demonstrate our approach in Sec.~\ref{sec:geo} exhibits clear advantage on all metrics, demonstrating temporal geometry constraint enhance the joint optimization of articulated modeling. Notably, ``w/ random $\Delta t$" brings negative effects, as using other states for difference approximation introduces additional motion errors especially in early optimization.



\begin{table}[h]
\centering
\small

\caption{\textbf{Ablation of temporal geometry constraints}. We report the average results on ArtGS-Multi and 2 screw-joint objects. }
\label{tab:ablation_geo}

\resizebox{\linewidth}{!}{
\begin{tabular}{ccccccc}
\toprule
Setting &Axis Ang &Axis Pos &Part Motion &CD-s &CD-m &CD-w \\
\midrule
w/o $\mathcal{L}_{\text{geo}}$ &4.04 &16.08 &4.73 &4.42 &5.70 &5.06\\
w/o $\nabla N$ &3.61 &9.52 &3.04 &3.86 &3.72 &4.41  \\
w/ random $\Delta t$ &5.83 &23.10 &6.24 &6.60 &10.07 &6.34\\
w/ tiny $\Delta t$ &2.65 &11.84 &3.74 &4.78 &3.58 &4.55 \\
w/ ours & \best{0.41} &\best{1.18} &\best{0.22} &\best{1.41} &\best{2.38} &\best{2.13}\\
\bottomrule 
\end{tabular}
}
\vspace{-5pt}
\end{table}





\begin{figure}[h]
\centering
\vspace{-0.5em}
\includegraphics[width=\linewidth]{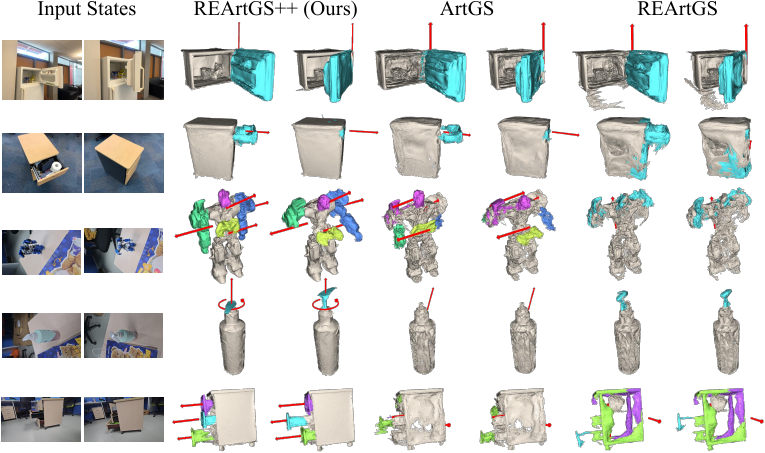}
\caption{Qualitative results of part-level surface reconstruction at both start and end states on real-world objects.}
\label{fig:real_world}
\vspace{-1.0em}
\end{figure}
\subsection{Generalization to Real-World Articulated Objects}
We present the quantitative results of 2 real-world objects from the PARIS dataset in Table.~\ref{tab:exp_paris}, and the qualitative results with additional 3 complex real objects we casually captured in Fig.~\ref{fig:real_world}. Experimental results prove that our method maintains superior articulated modeling results for multi-part and screw-joint objects in daily environments, demonstrating our generalization to the real world.

\section{Conclusion and Limitations}
In this paper, we propose REArtGS++, a novel method towards generalizable articulated object reconstruction via part-aware planar Gaussians. We model decoupled screw joint parameters to represent arbitrary motion and propose a time-continuous geometric regularization to enhance dynamic reconstruction. Both synthetic and real data experiments demonstrate our superior to previous SOTAs.

Our main limitations lie in transparent object reconstruction and the requirement for camera pose alignment between two states. Future work will combine depth estimation and transparent material modeling methods to address transparent surface perception~\cite{depthanything, li2025tsgs}, and jointly refine two-state camera poses~\cite{hong2024pf3plat} in real world data.

{
    \small
    \bibliographystyle{ieeenat_fullname}
    \bibliography{main}

@String(CVPR= {IEEE Conf. Comput. Vis. Pattern Recog.})

@String(ICCV= {Int. Conf. Comput. Vis.})

@String(ICLR = {Int. Conf. Learn. Represent.})

@String(CVPR  = {CVPR})

@String(ICCV  = {ICCV})

@String(ICLR  = {ICLR})

@inproceedings{wu2025reartgs,
  title={Reartgs: Reconstructing and generating articulated objects via 3d gaussian splatting with geometric and motion constraints},
  author={Wu, Di and Liu, Liu and Linli, Zhou and Huang, Anran and Song, Liangtu and Yu, Qiaojun and Wu, Qi and Lu, Cewu},
  booktitle={The Thirty-ninth Annual Conference on Neural Information Processing Systems},
  year={2025}
}

@INPROCEEDINGS{ASDF,
  author={Mu, Jiteng and Qiu, Weichao and Kortylewski, Adam and Yuille, Alan and Vasconcelos, Nuno and Wang, Xiaolong},
  booktitle={2021 IEEE/CVF International Conference on Computer Vision (ICCV)}, 
  title={A-SDF: Learning Disentangled Signed Distance Functions for Articulated Shape Representation}, 
  year={2021},
  volume={},
  number={},
  pages={12981-12991},
  doi={10.1109/ICCV48922.2021.01276}}

@ARTICLE{pgsr,
  author={Chen, Danpeng and Li, Hai and Ye, Weicai and Wang, Yifan and Xie, Weijian and Zhai, Shangjin and Wang, Nan and Liu, Haomin and Bao, Hujun and Zhang, Guofeng},
  journal={IEEE Transactions on Visualization and Computer Graphics}, 
  title={PGSR: Planar-Based Gaussian Splatting for Efficient and High-Fidelity Surface Reconstruction}, 
  year={2025},
  volume={31},
  number={9},
  pages={6100-6111},
  doi={10.1109/TVCG.2024.3494046}}

@inproceedings{jiang2022ditto,
   title={Ditto: Building Digital Twins of Articulated Objects from Interaction},
   author={Jiang, Zhenyu and Hsu, Cheng-Chun and Zhu, Yuke},
   booktitle={arXiv preprint arXiv:2202.08227},
   year={2022}
}

@inproceedings{liu2025building,
  title={Building Interactable Replicas of Complex Articulated Objects via Gaussian Splatting},
  author={Liu, Yu and Jia, Baoxiong and Lu, Ruijie and Ni, Junfeng and Zhu, Song-Chun and Huang, Siyuan},
  booktitle={The Thirteenth International Conference on Learning Representations},
  year={2025},
}

@inproceedings{jiayi2023paris,
  title={PARIS: Part-level Reconstruction and Motion Analysis for Articulated Objects},
  author={Liu, Jiayi and Mahdavi-Amiri, Ali and Savva, Manolis},
  booktitle={Proceedings of the IEEE/CVF International Conference on Computer Vision},
  pages={352--363},
  year={2023}
}

@Article{kerbl3Dgaussians,
      author       = {Kerbl, Bernhard and Kopanas, Georgios and Leimk{\"u}hler, Thomas and Drettakis, George},
      title        = {3D Gaussian Splatting for Real-Time Radiance Field Rendering},
      journal      = {ACM Transactions on Graphics},
      number       = {4},
      volume       = {42},
      month        = {July},
      year         = {2023},
      url          = {https://repo-sam.inria.fr/fungraph/3d-gaussian-splatting/}
}

@inproceedings{song2024reacto,
  title={REACTO: Reconstructing Articulated Objects from a Single Video},
  author={Song, Chaoyue and Wei, Jiacheng and Foo, Chuan Sheng and Lin, Guosheng and Liu, Fayao},
  booktitle={Proceedings of the IEEE/CVF Conference on Computer Vision and Pattern Recognition},
  pages={5384--5395},
  year={2024}
}

@inproceedings{weng2024neural,
      title={Neural Implicit Representation for Building Digital Twins of Unknown Articulated Objects}, 
      author={Yijia Weng and Bowen Wen and Jonathan Tremblay and Valts Blukis and Dieter Fox and Leonidas Guibas and Stan Birchfield},
      booktitle = {CVPR},
      year      = {2024},
    }

@inproceedings{Huang2DGS2024,
    title={2D Gaussian Splatting for Geometrically Accurate Radiance Fields},
    author={Huang, Binbin and Yu, Zehao and Chen, Anpei and Geiger, Andreas and Gao, Shenghua},
    publisher = {Association for Computing Machinery},
    booktitle = {SIGGRAPH 2024 Conference Papers},
    year      = {2024},
    doi       = {10.1145/3641519.3657428}
}

@article{yu2024gsdf,
  title={Gsdf: 3dgs meets sdf for improved rendering and reconstruction},
  author={Yu, Mulin and Lu, Tao and Xu, Linning and Jiang, Lihan and Xiangli, Yuanbo and Dai, Bo},
  journal={arXiv preprint arXiv:2403.16964},
  year={2024}
}

@inproceedings{scaffoldgs,
  title={Scaffold-gs: Structured 3d gaussians for view-adaptive rendering},
  author={Lu, Tao and Yu, Mulin and Xu, Linning and Xiangli, Yuanbo and Wang, Limin and Lin, Dahua and Dai, Bo},
  booktitle={Proceedings of the IEEE/CVF Conference on Computer Vision and Pattern Recognition},
  pages={20654--20664},
  year={2024}
}

@InProceedings{4dgs,
    author    = {Wu, Guanjun and Yi, Taoran and Fang, Jiemin and Xie, Lingxi and Zhang, Xiaopeng and Wei, Wei and Liu, Wenyu and Tian, Qi and Wang, Xinggang},
    title     = {4D Gaussian Splatting for Real-Time Dynamic Scene Rendering},
    booktitle = {Proceedings of the IEEE/CVF Conference on Computer Vision and Pattern Recognition (CVPR)},
    month     = {June},
    year      = {2024},
    pages     = {20310-20320}
}

@inproceedings{yang2023gs4d,
  title={Real-time Photorealistic Dynamic Scene Representation and Rendering with 4D Gaussian Splatting},
  author={Yang, Zeyu and Yang, Hongye and Pan, Zijie and Zhang, Li},
  booktitle={International Conference on Learning Representations (ICLR)},
  year={2024}
}

@article{huang2023sc,
  title={SC-GS: Sparse-Controlled Gaussian Splatting for Editable Dynamic Scenes},
  author={Huang, Yi-Hua and Sun, Yang-Tian and Yang, Ziyi and Lyu, Xiaoyang and Cao, Yan-Pei and Qi, Xiaojuan},
  journal={arXiv preprint arXiv:2312.14937},
  year={2023}
}

@inproceedings{
    spgs,
    title={Superpoint Gaussian Splatting for Real-Time High-Fidelity Dynamic Scene Reconstruction},
    author={Diwen Wan, Ruijie Lu, Gang Zeng},
    booktitle={Forty-first International Conference on Machine Learning},
    year={2024}
}

@inproceedings{shi2024language,
  title={Language embedded 3d gaussians for open-vocabulary scene understanding},
  author={Shi, Jin-Chuan and Wang, Miao and Duan, Hao-Bin and Guan, Shao-Hua},
  booktitle={Proceedings of the IEEE/CVF Conference on Computer Vision and Pattern Recognition},
  pages={5333--5343},
  year={2024}
}

@article{qin2023langsplat,
  title={LangSplat: 3D Language Gaussian Splatting},
  author={Qin, Minghan and Li, Wanhua and Zhou, Jiawei and Wang, Haoqian and Pfister, Hanspeter},
  journal={arXiv preprint arXiv:2312.16084},
  year={2023}
}

@article{shorinwa2024fast,
  title={Fast-splat: Fast, ambiguity-free semantics transfer in Gaussian splatting},
  author={Shorinwa, Ola and Sun, Jiankai and Schwager, Mac},
  journal={arXiv preprint arXiv:2411.13753},
  year={2024}
}

@article{kirillov2023segany,
  title={Segment Anything},
  author={Kirillov, Alexander and Mintun, Eric and Ravi, Nikhila and Mao, Hanzi and Rolland, Chloe and Gustafson, Laura and Xiao, Tete and Whitehead, Spencer and Berg, Alexander C. and Lo, Wan-Yen and Doll{\'a}r, Piotr and Girshick, Ross},
  journal={arXiv:2304.02643},
  year={2023}
}

@misc{zhang2022dino,
      title={DINO: DETR with Improved DeNoising Anchor Boxes for End-to-End Object Detection}, 
      author={Hao Zhang and Feng Li and Shilong Liu and Lei Zhang and Hang Su and Jun Zhu and Lionel M. Ni and Heung-Yeung Shum},
      year={2022},
      eprint={2203.03605},
      archivePrefix={arXiv},
      primaryClass={cs.CV}
}

@InProceedings{votesplat,
    author    = {Jiang, Minchao and Jia, Shunyu and Gu, Jiaming and Lu, Xiaoyuan and Zhu, Guangming and Dong, Anqi and Zhang, Liang},
    title     = {VoteSplat: Hough Voting Gaussian Splatting for 3D Scene Understanding},
    booktitle = {Proceedings of the IEEE/CVF International Conference on Computer Vision (ICCV)},
    month     = {October},
    year      = {2025},
    pages     = {6456-6465}
}

@article{guo2025articulatedgs,
  title={ArticulatedGS: Self-supervised Digital Twin Modeling of Articulated Objects using 3D Gaussian Splatting},
  author={Guo, Junfu and Xin, Yu and Liu, Gaoyi and Xu, Kai and Liu, Ligang and Hu, Ruizhen},
  journal={arXiv preprint arXiv:2503.08135},
  year={2025}
}

@article{shen2025gaussianart,
  title={GaussianArt: Unified Modeling of Geometry and Motion for Articulated Objects},
  author={Shen, Licheng and Zhang, Saining and Li, Honghan and Yang, Peilin and Huang, Zihao and Zhang, Zongzheng and Zhao, Hao},
  journal={arXiv preprint arXiv:2508.14891},
  year={2025}
}

@InProceedings{SAPIEN,
author = {Xiang, Fanbo and Qin, Yuzhe and Mo, Kaichun and Xia, Yikuan and Zhu, Hao and Liu, Fangchen and Liu, Minghua and Jiang, Hanxiao and Yuan, Yifu and Wang, He and Yi, Li and Chang, Angel X. and Guibas, Leonidas J. and Su, Hao},
title = {{SAPIEN}: A SimulAted Part-based Interactive ENvironment},
booktitle = {The IEEE Conference on Computer Vision and Pattern Recognition (CVPR)},
month = {June},
year = {2020}}

@inproceedings{wang2025adamanip,
    title={AdaManip: Adaptive Articulated Object Manipulation Environments and Policy Learning},
    author={Wang, Yuanfei and Zhang, Xiaojie and Wu, Ruihai and Li, Yu and Shen, Yan and Wu, Mingdong and He, Zhaofeng and Wang, Yizhou and Dong, Hao},
    booktitle={International Conference on Learning Representations},
    year={2025},
    url={https://openreview.net/forum?id=Luss2sa0vc}
  }

@article{Yu2025ArtGS3DGS,
  title={ArtGS:3D Gaussian Splatting for Interactive Visual-Physical Modeling and Manipulation of Articulated Objects},
  author={Qiaojun Yu and Xibin Yuan and Yu Jiang and Junting Chen and Dongzhe Zheng and Ce Hao and Yang You and Yixing Chen and Yao Mu and Liu Liu and Cewu Lu},
  journal={ArXiv},
  year={2025},
  volume={abs/2507.02600},
  url={https://api.semanticscholar.org/CorpusID:280126077}
}

@article{yu2024gamma,
  title={GAMMA: Generalizable Articulation Modeling and Manipulation for Articulated Objects},
  author={Yu, Qiaojun and Wang, Junbo and Liu, Wenhai and Hao, Ce and Liu, Liu and Shao, Lin and Wang, Weiming and Lu, Cewu},
  booktitle={2024 International Conference on Robotics and Automation (ICRA)},
  year={2024},
  organization={IEEE},
}

@misc{li2025tsgs,
  title={TSGS: Improving Gaussian Splatting for Transparent Surface Reconstruction via Normal and De-lighting Priors}, 
  author={Mingwei Li and Pu Pang and Hehe Fan and Hua Huang and Yi Yang},
  year={2025},
  eprint={2504.12799},
  archivePrefix={arXiv},
  primaryClass={cs.CV},
  url={https://arxiv.org/abs/2504.12799}, 
}

@inproceedings{depthanything,
      title={Depth Anything: Unleashing the Power of Large-Scale Unlabeled Data}, 
      author={Yang, Lihe and Kang, Bingyi and Huang, Zilong and Xu, Xiaogang and Feng, Jiashi and Zhao, Hengshuang},
      booktitle={CVPR},
      year={2024}
}

@article{hong2024pf3plat,
      title   = {PF3plat: Pose-Free Feed-Forward 3D Gaussian Splatting},
      author  = {Sunghwan Hong and Jaewoo Jung and Heeseong Shin and Jisang Han and Jiaolong Yang and Chong Luo and Seungryong Kim},
      journal = {arXiv preprint arXiv:2410.22128},
      year    = {2024}
    }

@inproceedings{zhang2025cob,
  title={Cob-gs: Clear object boundaries in 3dgs segmentation based on boundary-adaptive gaussian splitting},
  author={Zhang, Jiaxin and Jiang, Junjun and Chen, Youyu and Jiang, Kui and Liu, Xianming},
  booktitle={Proceedings of the Computer Vision and Pattern Recognition Conference},
  pages={19335--19344},
  year={2025}
}
}


\end{document}